%% file: tplp.tex
\documentclass{tlp}

\usepackage[T1]{fontenc}
\usepackage{amsmath,amssymb}
\usepackage{epsfig}

\newcommand{\wrt}{{\it w.r.t. }}    
\newcommand{\eg}{\emph{e.g., }}     
\newcommand{\ie}{\emph{i.e.} }      

\newcommand{\setcomp}[1]{\widetilde{#1}}

\newcommand{\fig}{Fig. }

\newtheorem{definition}{Definition}
\newtheorem{example}{Example}
\newtheorem{lemma}{Lemma}
\newtheorem{theorem}{Theorem}

\newtheorem{proposition}{Proposition}
\newtheorem{remark}{Remark}

\submitted{6 June 2007}

\revised{17 December 2007}

\accepted{29 February 2008}

\begin{document}

\title{Preferred extensions as stable models}

\title[Preferred extensions as stable models]
{Preferred extensions as stable models
\thanks{This is a revised and improved version of
the paper \emph{Inferring preferred extensions by minimal models}
which appeared in Guillermo R. Simari and Paolo Torroni (Eds),
proceedings of the workshop Argumentation and Non-Monotonic
Reasoning (LPNMR-07 Workshop).}}

\author[J. C. Nieves, M. Osorio, and U. Cort\'es]
{ JUAN CARLOS NIEVES, ULISES CORT\'ES  \\
Universitat Polit\`ecnica de Catalunya\\
Software Department (LSI)\\
c/Jordi Girona 1-3, E08034, Barcelona, Spain  \\
\email{\{jcnieves,ia\}@lsi.upc.edu} \and
MAURICIO OSORIO \\
Universidad de las Am\'ericas - Puebla\\
CENTIA \\
Sta. Catarina M\'artir, Cholula, Puebla, 72820 M\'exico\\
\email{osoriomauri@googlemail.com} }

\maketitle

\begin{abstract}

Given an argumentation framework \emph{AF}, we introduce a mapping
function that constructs a disjunctive logic program \emph{P}, such
that the preferred extensions of \emph{AF} correspond to the stable
models of \emph{P}, after intersecting each stable model with the
relevant atoms. The given mapping function is of polynomial size
\wrt \emph{AF}.

In particular, we identify that there is a direct relationship
between the minimal models of a propositional formula and the
preferred extensions of an argumentation framework by working on
representing the defeated arguments. Then we show how to infer the
preferred extensions of an argumentation framework by using UNSAT
algorithms and disjunctive stable model solvers. The relevance of
this result is that we define a direct relationship between one of
the most satisfactory argumentation semantics and one of the most
successful approach of non-monotonic reasoning \ie logic programming
with the stable model semantics.

\end{abstract}


\section{Introduction}\label{groundedSem}

Dung's approach, presented in \cite{Dung95}, is a unifying framework
which has played an influential role on argumentation research and
Artificial Intelligence (AI). In fact, Dung's approach has
influenced subsequent proposals for argumentation systems, \eg
\cite{Ben02}. Besides, Dung's approach is mainly relevant in fields
where conflict management plays a central role. For instance, Dung
showed that his theory naturally captures the solutions of the
theory of n-person games and the well-known stable marriage problem.

Dung defined four argumentation semantics: \emph{stable semantics},
\emph{preferred semantics}, \emph{grounded semantics}, and
\emph{complete semantics}. The central notion of these semantics is
the \emph{acceptability of the arguments}.
The main argumentation semantics for collective acceptability are
the grounded semantics and the preferred semantics
\cite{PV02,aspic:D2.2}. The first one represents a skeptical
approach and the second one represents a credulous approach.

Dung showed that argumentation can be viewed as logic programming
with \emph{negation as failure}. Specially, he showed that the
grounded semantics can be characterized by the well-founded
semantics \cite{GelderRS91}, and the stable semantics by the stable
model semantics \cite{GelLif91}. This result is of great importance
because it introduces a general method for generating
metainterpreters for argumentation systems \cite{Dung95}. Following
this issue, we will prove that it is possible to characterize the
preferred semantics based on the minimal models of a propositional
formula (Theorem \ref{theo:mimModelPref}). We will also show that
the preferred semantics can be characterized by the stable models of
a positive disjunctive logic program (Theorem \ref{prefe-General}).
The importance of this characterization is that we are defining a
direct relationship between one of the most satisfactory
argumentation semantics and may be the most successful approach of
non-monotonic reasoning of the last two decades \ie logic
programming with the stable model semantics.

As a natural consequence of our result, we present two easy-to-use
forms for inferring the preferred extensions of an argumentation
framework (\emph{AF}). The first one is based on a mapping function
which is quadratic size \wrt the number of arguments of \emph{AF}
and UNSAT algorithms. The second one is also based on a mapping
function which is quadratic size \wrt the number of arguments of
\emph{AF} and disjunctive stable model solvers.

It is worth mentioning that the decision problem of the preferred
semantics is hard since it is co-NP-Complete \cite{DunBen04}. In
fact, we can find different strategies for computing the preferred
semantics \cite{BesDou04,CDM03,DunKowTon06,DunManTon07}. However, we
can find really few implementations of them
\cite{aspic:arg-engine,GaeTon07}. One of the relevant points of our
result is that we can take advance of efficient disjunctive stable
model solvers, \eg the DLV System \cite{DLV}, for inferring the
preferred semantics. The DLV System is a successful stable model
solver that includes deductive database optimization techniques, and
non-monotonic reasoning optimization techniques in order to improve
its performance \cite{LeonePFCDEGIIKPP02,GebserLNNST07}. In fact, we
can implement the preferred semantics inside object-oriented
programs based on our characterization and the DLV JAVA Wrapper
\cite{Ricca03}.

The rest of the paper is divided as follows:  In \S \ref{sec:back},
we present some basic concepts of logic programs and argumentation
theory. In \S \ref{pref-UNSAT}, we present a characterization of the
preferred semantics by minimal models. In \S
\ref{preferGeneralProg}, we present how to compute the preferred
semantics by using the minimal models of a positive disjunctive
logic program. Finally in the last section, we present our
conclusions.

\section{Background}\label{sec:back}
In this section, we present the syntax of a valid logic program, the
definition of the stable model semantics, and the definition of the
preferred semantics. We will use basic well-known definitions in
complexity theory such as that of co-NP-complete problem.

\subsection{Logic Programs: Syntax}\label{backSyntax}
The language of a propositional logic has an alphabet consisting of
\begin{description}
\item[(i)] A signature ${\cal L}$ that is a finite set of elements that we call
atoms, denoted usually as $p_0, p_1, ...$
\item[(ii)] connectives : $\vee , \wedge , \leftarrow , \lnot , \bot, \top$
\item[(iii)] auxiliary symbols : ( , ).
\end{description}
where $\vee , \wedge , \leftarrow$ are 2-place connectives, $\lnot$
is 1-place connective and $\bot, \top$ are 0-place connectives or
constant symbols.  A literal is an atom, $a$, or the negation of an
atom $\lnot a$. Given a set of atoms $\{a_{1},...,a_{n}\}$, we write
$\lnot \{a_{1},...,a_{n}\}$ to denote the set of literals $\{\lnot
a_{1},...,\lnot a_{n}\}.$ Formul\ae\ are constructed as usual in
logic. A theory $T$ is a finite set of formul\ae. By ${\cal L}_T$,
we denote the signature of \emph{T}, namely the set of atoms that
occur in \emph{T}.

A general clause, {\it C}, is denoted by $a_1 \vee \ldots \vee a_m$
$\leftarrow$ $l_1, \dots, l_n$,\footnote{ $l_1, \dots, l_n$
represents the formula $l_1 \wedge \dots \wedge l_n$.} where $m \geq
0$, $n \geq 0$, $m+n > 0$, each $a_i$ is an atom, and each $l_i$ is
a literal. When $n=0$ and $m>0$ the clause is an abbreviation of
$a_1\vee
\ldots \vee a_m \leftarrow \top$. 
When $m=0$ the clause is an abbreviation of $\bot \leftarrow l_1
,\dots,l_n$. Clauses of this form are called constraints (the rest,
non-constraint clauses). A general program, $P$, is a finite set of
general clauses.  Given a universe $U$, we define the
\emph{complement} of a set $S \subseteq U$ as $\setcomp{S} = U
\setminus S$.

We point out that whenever we consider logic programs our negation
$\lnot$ corresponds to the default negation $not$ used in Logic
Programming. Also, it is convenient to remark that in this paper we
are not using at all the so called \emph{strong negation} used in
ASP.

\subsection{Stable Model Semantics}
First, to define the stable model semantics, let us define some
relevant concepts.

\begin{definition} Let $T$ be a theory, an interpretation $I$ is a mapping from
${\cal L}_T$ to $\{0, 1\}$ meeting the conditions:
\begin{enumerate}
\item $I(a \wedge b) = min\{I(a), I(b)\}$,
\item $I(a \vee b) = max\{I(a), I(b)\}$,
\item $I(a\leftarrow b) = 0$ iff $I(b) = 1$ and $I(a) = 0$,
\item $I(\lnot a) = 1 - I(a)$,
\item $I(\bot) = 0$.
\item $I(\top)=1$.
\end{enumerate}
\end{definition}

It is standard to provide interpretations only in terms of a mapping
from ${\cal L}_T$ to $\{0, 1\}$. Moreover it is easy to prove that
this mapping is unique by virtue of the definition by recursion
\cite{Dalen94}.

An interpretation $I$ is called a model of $P$ iff for each clause
$c \in P$, $I(c) = 1$. A theory is consistent if it admits a model,
otherwise it is called inconsistent. Given a theory $T$ and a
formula $\alpha$, we say that $\alpha$ is a logical consequence of
$T$, denoted by $T \models \alpha$, if for every model $I$ of $T$ it
holds that $I(\alpha)=1$. It is a well known result that $T \models
\alpha$ iff $T \cup \{\neg \alpha \}$ is inconsistent. It is
possible to identify an interpretation with a subset of a given
signature. For any interpretation, the corresponding subset of the
signature is the set of all atoms that are true  \wrt the
interpretation. Conversely, given an arbitrary subset of the
signature, there is a corresponding interpretation defined by
specifying  that the mapping assigned to an atom in the subset is
equal to 1 and otherwise to 0. We use this view of interpretations
freely in the rest of the paper.

We say that a model $I$ of a theory $T$  is a minimal model  if
there does not exist a model $I'$ of $T$ different from $I$ such
that $I' \subset I$.
Maximal models are defined in the analogous form.

By using logic programming with stable model semantics, it is
possible to describe a computational problem as a logic program
whose stable models correspond to the solutions of the given
problem.
The following definition of a stable model for general
programs 
was presented in \cite{GelLif91}.

Let \emph{P} be any general program. For any set $S \subseteq {\cal
L}_P$, let $P^S$ be the general program obtained from \emph{P} by
deleting

\begin{description}
\item[(i)] each rule that has a formula $\lnot l$ in its body with
$l \in S$, and then

\item[(ii)] all formul\ae\ of the form $\lnot l$ in the bodies of
the remaining rules.
\end{description}

\noindent Clearly $P^S$ does not contain $\lnot$. Hence \emph{S} is
a stable model of \emph{P} iff \emph{S} is a minimal model of $P^S$.

In order to illustrate this definition let us consider the following
example:

\begin{example}
Let $S = \{ b \}$ and $P$ be the following logic program:

\begin{tabular}{ll}
  $b \leftarrow \neg a$. $~~~~~~~~~~~~~~~~~$ & $b \leftarrow \top$. \\
  $c \leftarrow \neg b$. $~~~~~~~~~~~~~~~~~$ & $c \leftarrow a$. \\
\end{tabular}

\noindent We can see that $P^S$ is:

\begin{tabular}{ll}
  $b \leftarrow \top$. $~~~~~~~~~~~~~~~~~$ & $c \leftarrow a$. \\
\end{tabular}

\noindent Notice that $P^S$ has two models: $\{b\}$ and $\{a,
b,c\}$. Since the minimal model amongst these models is $\{b\}$, we
can say that $S$ is a stable model of $P$.

\end{example}

\subsection{Argumentation theory}

Now, we define some basic concepts of Dung's argumentation approach.
The first one is that of an argumentation framework. An
argumentation framework captures the relationships between the
arguments (All the definitions of this subsection were taken from
the seminal paper \cite{Dung95}).

\begin{definition}
An argumentation framework is a pair $AF = \langle AR,
attacks\rangle$, where \emph{AR} is a finite set of arguments, and
\emph{attacks} is a binary relation on \emph{AR}, \ie \emph{attacks}
$\subseteq AR \times AR$.
\end{definition}


For two arguments $a$ and $b$, we say that $a$ \emph{attacks} $b$
(or $b$ is attacked by $a$) if $attacks(a,b)$ holds. Notice that the
relation \emph{attacks} does not yet tell us with which arguments a
dispute can be won; it only tells us the relation of two conflicting
arguments.

\begin{figure}[th]\label{}
\begin{center}
\epsfig{file=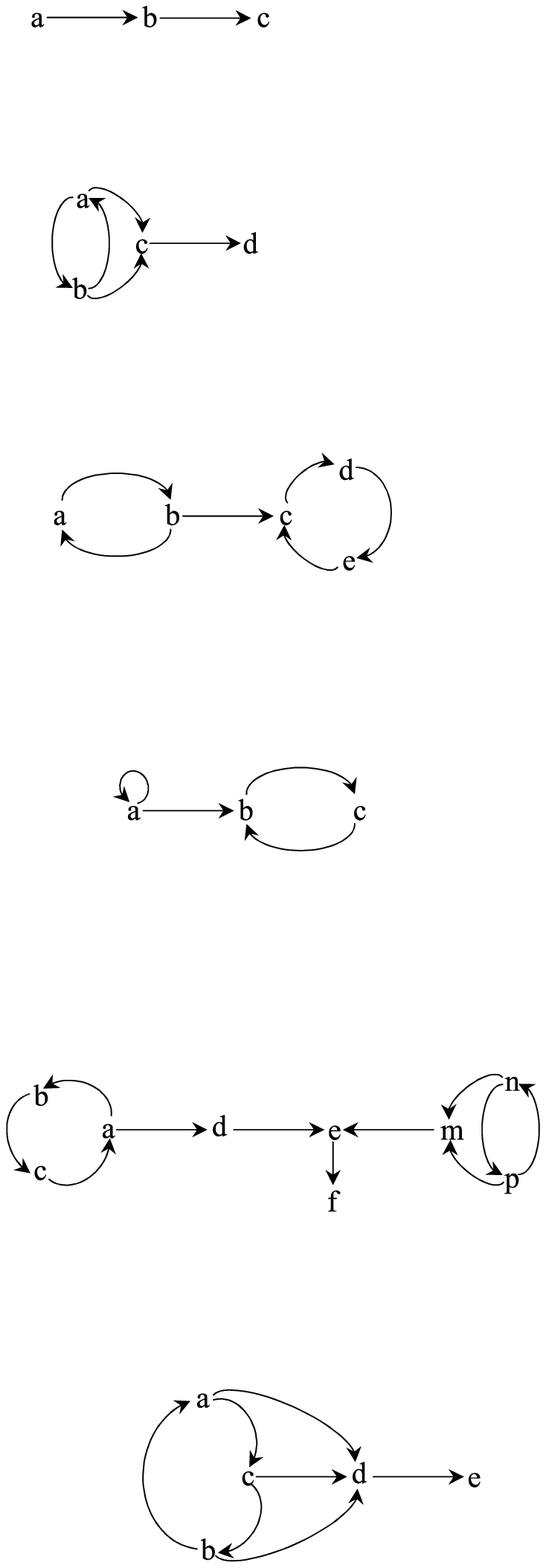, scale=1.0,bbllx=80, bblly=715, bburx=192,
bbury=740, clip=} \caption{Graph representation of the argumentation
framework $AF = \langle \{a ,b, c \}, \{(a,b), (b,c) \} \rangle
$.}\label{fig1}
\end{center}
\end{figure}

It is worth mentioning that any argumentation framework can be
regarded as a directed graph. For instance, if $AF = \langle
\{a,b,c\}, \{ (a,b), (b,c) \} \rangle$, then $AF$ can be represented
as shown in \fig \ref{fig1}.


\begin{definition}\label{confFree}
A set {\it S} of arguments is said to be conflict-free if there are
no arguments {\it a, b} in {\it S} such that {\it a attacks b}.
\end{definition}

A central notion of Dung's framework is \emph{acceptability}. It
captures how an argument that cannot defend itself, can be protected
by a set of arguments.

\begin{definition}
(1) An argument $a \in AR$ is \emph{acceptable} \wrt a set $S$ of
arguments iff for each argument $b \in AR$: If $b$ attacks $a$ then
$b$ is attacked by an argument in $S$. (2) A conflict-free set of
arguments $S$ is \emph{admissible} iff each argument in $S$ is
acceptable \wrt $S$.
\end{definition}

Let us consider the argumentation framework $AF$ of \fig \ref{fig1}.
We can see that
$AF$ has three admissible sets: $\{\}$, $\{a\}$ and $\{a, c\}$.
Intuitively, an admissible set is a coherent point of view. Since an
argumentation framework could have several coherent point of views,
one can take the maximum admissible sets in order to get maximum
coherent point of views of an argumentation framework. This idea is
captured by Dung's framework with the concept of \emph{preferred
extension}.

\begin{definition}
A preferred extension of an argumentation framework $AF$ is a maximal (\wrt inclusion) admissible set of $AF$.
\end{definition}

Since an argumentation framework could have more than one preferred
extension, the preferred semantics is called credulous. The
argumentation framework of \fig \ref{fig1} has just one preferred
extension which is $\{a,c\}$.

\begin{remark}
By definition, it is clear that any argument which belongs to a
preferred extension $E$ is acceptable \wrt $E$. Hence we will say
that any argument which does not belong to some preferred extension
is a \emph{defeated argument}.
\end{remark}

\input{unsat}
\input{prefExtGenProg}

\section{Conclusions}

Since Dung introduced his abstract argumentation approach, he proved
that his approach can be regarded as a special form of logic
programming with \emph{negation as failure}. In fact, he showed the
grounded and stable semantics can be characterized by the
well-founded and stable models semantics respectively.  This result
is important because it defined a general method for generating
metainterpreters for argumentation systems \cite{Dung95}. Concerning
this issue, Dung did not give any characterization of the preferred
semantics in terms of logic programming semantics. It is worth
mentioning that according to the literature
\cite{PV02,aspic:D2.2,Pollock95,BonDunKowTon97,Dung95}, the
preferred semantics is regarded as one of the most satisfactory
argumentation semantics of Dung's argumentation approach.

In this paper, we characterize the preferred semantics in terms of
minimal models (see Theorem \ref{theo:mimModelPref}) and stable
model semantics (see Theorem \ref{prefe-General}). These
characterizations are based on two mapping functions that construct
a propositional formula and a disjunctive logic program
respectively. These characterizations have as main result the
definition of a direct relationship between one of the most
satisfactory argumentation semantics and may be the most successful
approach of non-monotonic reasoning of the last two decades \ie
logic programming with the stable model semantics. Based on this
fact, we introduce a novel and easy-to-use method for implementing
argumentation systems which are based on the preferred semantics. It
is quite obvious that our method will take advantage of the platform
that has been developed under stable model semantics for generating
argumentation systems. For instance, we can implement the preferred
semantics inside object-oriented programs based on our
characterization (Theorem \ref{prefe-General}, Proposition
\ref{lemma:prefDisjNF}) and the DLV JAVA Wrapper \cite{Ricca03}.




We can see that our approach falls in the family of the
model-checking methods for inferring the preferred semantics. In
fact, our approach is closely related to the methods suggested in
\cite{BesDou04,EglWol06}. As seen in Theorem
\ref{theo:mimModelPref}, our propositional formula $\alpha(AF)$ is
closely related to one of the propositional formul\ae\ (see
Proposition \ref{propBesDou04}) which were suggested in
\cite{BesDou04}. It is worth mentioning that the propositional
formula suggested by \cite{EglWol06} for inferring the admissible
sets of an argumentation framework is the same to the propositional
formula of Proposition \ref{propBesDou04}. The main difference
between the approaches suggested by \cite{BesDou04,EglWol06} and our
approach is the strategy for inferring the models of a propositional
formula. Instead of using \emph{maximal models} for characterizing
the preferred semantics as it is done dy \cite{BesDou04}, we are
using \emph{ minimal models/stable models}. Hence, we can use any
system which could compute minimal models/stable models of a
propositional formula. Maximality in Egly and Woltran' approach is
checked on the object level, \ie within the resulting Quantified
Boolean formula (QBF).

An interesting property of our approach is that whenever we use
stable model solvers for computing the preferred extensions of an
argumentation framework, we can compute all the preferred extensions
in full. In decision-making systems, it is not strange to require
all the possible coherent points of view (preferred extensions) in a
dispute between arguments. For instance, in the medical domain when
a doctor has to give a diagnosis under incomplete information, he
has to consider all the possible alternatives in his decisions
\cite{CTNLC05,TCNLC05}.



\section*{Acknowledgement}
We are grateful to anonymous referees for their useful comments.
J.C. Nieves thanks to CONACyT for his PhD Grant. J.C. Nieves and U.
Cort\'es would like to acknowledge support from the EC funded
project SHARE-it: Supported Human Autonomy for Recovery and
Enhancement of cognitive and motor abilities using information
technologies (FP6-IST-045088). The views expressed in this paper are
not necessarily those of the SHARE-it consortium.

\bibliographystyle{acmtrans}
\bibliography{../../../biblio/papers_jcns}

\input{proofs}

\end{document}

%% file: unsat.tex
\section{Preferred extensions and UNSAT problem}\label{pref-UNSAT}

In this section, we will define a mapping function that constructs a
propositional formula, such that its minimal models characterize the
preferred extensions of an argumentation framework. This
characterization will provide a method for computing preferred
extensions based on Model Checking and Unsatisfiability (UNSAT).


In order to characterize the preferred semantics in terms of minimal models, we will introduce some concepts.

\begin{definition}
Let $T$ be a theory with signature ${\cal L}$. We say that ${\cal
L^\prime}$ is a copy-signature of ${\cal L}$ iff

\begin{itemize}
  \item ${\cal L}\cap{\cal L^\prime}=\emptyset$,
  \item the cardinality of ${\cal L^\prime}$ is the same to ${\cal
  L}$ and
  \item there is a bijective function $f$ from ${\cal L}$ to ${\cal
L^\prime}$.
\end{itemize}
\end{definition}

It is well known that there exists a bijective function from one set
to another if both sets have the same cardinality.  Now one can
establish an important relationship between maximal and
minimal models. 

\begin{proposition}\label{pseudo-dualidad}
Let $T$ be a theory with signature ${\cal L}_T$. Let ${\cal
L^\prime}$ be a copy-signature of ${\cal L}_T$.
 By $g(T)$ we denote the theory obtained from $T$ by replacing every
occurrence  of an atom $x$ in $T$ by $\neg f(x)$. Then $M$ is a
maximal model of $T$ iff $f({\cal L}_T \setminus M)$ is a minimal
model of $g(T)$.
\end{proposition}
\begin{proof}
See Appendix A.

\end{proof}


Our representations of an argumentation framework use the predicate
\emph{d(x)}, where the intended meaning of \emph{d(x)} is: ``the
argument \emph{x} is defeated''. By considering the predicate
$d(x)$, we will define a mapping function from an argumentation
framework to a propositional formula. This propositional formula
captures two basic conditions which make an argument to be defeated.

\begin{definition}\label{formulaUNSAT}
Let $AF  = \langle AR,attacks\rangle$ be an argumentation framework,
then $\alpha(AF)$ is defined as follows:

{\small
\[
\alpha(AF)  = \bigwedge_{a \in AR} ((\bigwedge_{b :(b,a) \in attacks
} d(a) \leftarrow \neg d(b)) \wedge ( \bigwedge_{b :(b,a) \in
attacks} d(a) \leftarrow \bigwedge_{c :(c,b) \in attacks} d(c) ))
\]}
\end{definition}


\begin{enumerate}
  \item The first condition of $\alpha(AF)$ $(\bigwedge_{b :(b,a) \in attacks} d(a) \leftarrow \neg d(b))$
suggests that the argument $a$ is defeated when any one of its
adversaries is not defeated.
  \item The second condition of $\alpha(AF)$ $( \bigwedge_{b :(b,a) \in attacks}  d(a)
\leftarrow \bigwedge_{c :(c,b) \in attacks} d(c) )$ suggests that
the argument $a$ is defeated when all the arguments that
defend\footnote{We say that $c$ defends $a$ if $b$ attacks $a$ and
$c$ attacks $b$.} $a$ are defeated.
\end{enumerate}

Since $\alpha(AF)$ captures conditions which make an argument to be
defeated, it is quite obvious that any argument which satisfies
these conditions could not belong to an admissible set. Therefore
these arguments also could not belong to a preferred extension.

Notice that $\alpha(AF)$ is a \emph{finite grounded formula}, this
means that it does not contain predicates with variables; hence,
$\alpha(AF)$ is essentially a propositional formula (just
considering the atoms like $d(a)$ as $d\_a$) of propositional logic.
In order to illustrate the propositional formula $\alpha(AF)$, let
us consider the following example.

\begin{example}\label{exam-alpha}
Let $AF  = \langle AR,attacks\rangle$ be the argumentation framework
of \fig \ref{fig1}. We can see that $\alpha(AF)$ is:








\[
\begin{array}{l}
( d(b) \leftarrow \neg d(a)) \wedge ( d(b) \leftarrow \top)  \wedge
(d(c) \leftarrow \neg d(b)) \wedge (d(c) \leftarrow d(a))
\end{array}
\]

\noindent Observe that $\alpha(AF)$ has no propositional clauses
\wrt argument $a$. This is essentially because $\alpha(AF)$ is
capturing the arguments which could be defeated and the argument $a$
will be always an acceptable argument.
\end{example}

It is worth mentioning that given an argumentation framework $AF$,
$\alpha(AF)$ will have at most $2n^2$ propositional clauses such
that $n$ is the number of arguments in $AR$ and the maximum
length\footnote{The length of our propositional clauses $C$ is given
by the number of atoms in the head of $C$ plus the number of
literals in the body of $C$} of each propositional clause is $n +
1$. Hence, we can say that $\alpha(AF)$ is quadratic size \wrt the
number of arguments of $AF$.

Essentially $\alpha(AF)$ is a propositional representation of the
argumentation framework $AF$. However $\alpha(AF)$ has the property
that its minimal models characterize $AF$'s preferred extensions. In
order to formalize this property, let us consider the following
proposition which was proved by Besnard and Doutre  in
\cite{BesDou04}.

\begin{proposition}\cite{BesDou04} \label{propBesDou04}
Let $AF = \langle AR,attacks\rangle$ be an argumentation framework.
Let $\beta(AF)$ be the formula:

\[
\bigwedge_{a \in AR} ((a \rightarrow \bigwedge_{b :(b,a) \in
attacks} \neg b) \wedge (a \rightarrow \bigwedge_{b :(b,a) \in
attacks} ( \;\;\;\; \bigvee_{c :(c,b) \in attacks} c )))
\]

\noindent then, a set $S \subseteq AR$ is a preferred extension iff
S is a maximal model of the formula $\beta(AF)$.
\end{proposition}

\noindent In contrast with $\alpha(AF)$ which captures conditions
which make an argument to be defeated, $\beta(AF)$ captures
conditions which make an argument acceptable. However, we will prove
that when the mapping $f(x)$ of the theory $g(\beta(AF))$
corresponds to $d(x)$ such that $x \in AF$, $\alpha(AF)$ is
logically equivalent to $g(\beta(AF))$ (see the proof of Theorem
\ref{theo:mimModelPref}). For instance, let us consider the
argumentation framework $AF$ of Example \ref{exam-alpha}. The
formula $\beta(AF)$ is:

\[
\begin{array}{l}
(\neg a \leftarrow b) \wedge (\bot \leftarrow b) \wedge (\neg b
\leftarrow c) \wedge (a \leftarrow c)
\end{array}
\]

\noindent If we replace each atom $x$ by the expression $\neg d(x)$, we get:

\[
\begin{array}{l}
(\neg \neg d(a) \leftarrow \neg d(b)) \wedge (\bot \leftarrow \neg
d(b)) \wedge (\neg \neg d(b) \leftarrow \neg d(c)) \wedge (\neg d(a)
\leftarrow \neg d(c))
\end{array}
\]

\noindent Now, if we apply transposition to each implication, we
obtain:

\[
\begin{array}{l}
(d(b) \leftarrow \neg d(a)) \wedge (d(b) \leftarrow \top) \wedge (
d(c) \leftarrow \neg d(b)) \wedge ( d(c) \leftarrow  d(a))
\end{array}
\]

\noindent  The latter formula corresponds to $\alpha (AF)$. The
following theorem is a straightforward consequence of Proposition
\ref{propBesDou04} and Proposition \ref{pseudo-dualidad}. Given  an
argumentation framework $AF = \langle AR, attacks \rangle $ and  $E
\subseteq AR$, we define the set $compl(E)$ as $\{d(a) | a \in AR
\setminus E\} $. Essentially, $compl(E)$ expresses the complement of
$E$ \wrt $AR$.

\begin{theorem}\label{theo:mimModelPref}
Let $AF = \langle AR,attacks\rangle$ be an argumentation framework
and $S \subseteq AR$. When the mapping $f(x)$ of the theory
$g(\beta(AF))$ corresponds to $d(x)$ such that $x \in AR$, the
following condition holds:
$S$ is a preferred extension of $AF$ iff $compl(S)$ is a minimal
model of $\alpha(AF)$.
\end{theorem}

\begin{proof}
See Appendix A.
\end{proof}

This theorem shows that it is possible to characterize the preferred
extensions of an argumentation framework $AF$ by considering the
minimal models of $\alpha(AF)$. In order to illustrate Theorem
\ref{theo:mimModelPref}, let us consider again $\alpha(AF)$ of
Example \ref{exam-alpha}. This formula has three models: $\{d(b)\}$,
$\{d(b), d(c)\}$ and $\{d(a), d(b), d(c)\}$. Then, the only minimal
model is $\{d(b)\}$, this implies that $\{a,c\}$ is the only
preferred extension of \emph{AF}. In fact, each model of
$\alpha(AF)$ implies an admissible set of \emph{AF}, this means that
$\{a,c\}$, $\{a\}$ and $\{\}$ are the admissible sets of \emph{AF}.

There is a well known relationship between minimal models and
logical consequence, see \cite{OsNaAr:tplp}. The following
proposition is a direct consequence of such relationship. Let $S$ be
a set of well formed formul\ae\ then we define $SetToFormula(S) =
\bigwedge_{c \in S} c$.

\begin{proposition}\label{lemmaConsequence}
Let $AF =\langle AR,attacks\rangle$ be an argumentation framework
and $S \subseteq AR$. $S$ is a preferred extension of $AF$ iff
$compl(S)$ is a model of $\alpha(AF)$ and $\alpha(AF) \wedge
SetToFormula( \lnot\setcomp{compl(S))} \models
SetToFormula(compl(S))$.
\end{proposition}
\begin{proof}
See Appendix A.
\end{proof}

There are several well-known approaches for inferring minimal models
from a propositional formula \cite{DimTor96,Ben05}. For instance, it
is possible to use UNSAT's algorithms for inferring minimal models.
Hence, it is clear that we can use UNSAT's algorithms for computing
the preferred extensions of an argumentation framework. This idea is
formalized with the following proposition.

\begin{theorem}\label{lemmaUNSAT}
Let $AF = \langle AR,attacks\rangle$ be an argumentation framework
and $S \subseteq AR$. S is a preferred extension of AF if and only
if $compl(S)$ is a model of $\alpha(AF)$ and $\alpha(AF) \wedge
SetToFormula( \lnot\setcomp{compl(S))} \wedge \lnot
SetToFormula(compl(S))$ is unsatisfiable.
\end{theorem}
\begin{proof}
Directly, by Proposition \ref{lemmaConsequence}.
\end{proof}

In order to illustrate Theorem \ref{lemmaUNSAT}, let us consider
again the argumentation framework $AF$ of Example \ref{exam-alpha}.
Let $S = \{a\}$, then $compl(S) = \{d(b),d(c)\}$. We have already
seen that $\{d(b),d(c)\}$ is a model of $\alpha(AF)$, hence the
formula to verify its unsatisfiability is:

\[
\begin{array}{l}
( d(b) \leftarrow \neg d(a)) \wedge ( d(b) \leftarrow \top)  \wedge
(d(c) \leftarrow \neg d(b)) \wedge (d(c) \leftarrow d(a)) \wedge \\
\neg d(a) \wedge (\neg d(b) \vee \neg d(c))
\end{array}
\]

\noindent However, this formula is satisfiable by the model
$\{d(b)\}$, then $\{a\}$ is not a preferred extension. Now, let $S
=\{a,c\}$, then $compl(S) = \{d(b)\}$. As seen before, $\{d(b)\}$ is
also a model of $\alpha(AF)$, hence the formula to verify its
unsatisfiability is:

\[
\begin{array}{l}
( d(b) \leftarrow \neg d(a)) \wedge ( d(b) \leftarrow \top)  \wedge
(d(c) \leftarrow \neg d(b)) \wedge (d(c) \leftarrow d(a)) \wedge \\
\neg d(a) \wedge \neg d(c) \wedge \neg d(b)
\end{array}
\]

\noindent It is easy to see that this formula is unsatisfiable, therefore $\{a,c\}$ is a preferred extension.

The relevance of Theorem \ref{lemmaUNSAT} is that UNSAT is the
prototypical and best-researched co-NP-complete problem. Hence,
Theorem \ref{lemmaUNSAT} opens the possibilities for using a wide
variety of algorithms for inferring the preferred semantics.





%% file: prefExtGenProg.tex
\section{Preferred extensions and general programs}\label{preferGeneralProg}

We have seen that the minimal models of $\alpha(AF)$ characterize
the preferred extensions of $AF$. One interesting point of
$\alpha(AF)$ is that $\alpha(AF)$ is logically equivalent to the
positive disjunctive logic program $\Gamma_{AF}$ (defined below). It
is well known that given a positive disjunctive logic program $P$,
all the minimal models of $P$ correspond to the stable models of
$P$. This property will be enough for characterizing the preferred
semantics by the stable models of the positive disjunctive logic
program $\Gamma_{AF}$.


We start this section by defining a mapping function which is a
variation of the mapping of Definition \ref{formulaUNSAT}.

\begin{definition}\label{def:mappingDisjunctive}
Let  $AF = \langle AR,attacks\rangle$ be an argumentation framework
and $a \in AR$. We define the transformation function $\Gamma(a)$ as
follows:

\[
\Gamma(a) = \{ \bigcup_{b :(b,a) \in attacks} \{ d(a) \vee d(b) \}
\} \cup \{ \bigcup_{b :(b,a) \in attacks}  \{ d(a) \leftarrow
\bigwedge_{c :(c,b) \in attacks} d(c) \} \}
\]
\end{definition}

Now we define the function $\Gamma$ in terms of an argumentation
framework.

\begin{definition}\label{pre-General-Prog}
Let  $AF = \langle AR,attacks\rangle$ be an argumentation framework.
We define its associated general program as follows:

\[
\Gamma_{AF} = \bigcup_{a \in AR} \Gamma(a)
\]
\end{definition}

\begin{remark}
Notice that $\alpha(AF)$ (see Definition  \ref{formulaUNSAT}) is
similar to $\Gamma_{AF}$. The main syntactic difference of
$\Gamma_{AF}$ \wrt $\alpha(AF)$ is the first part of $\Gamma_{AF}$
which is $(\bigwedge_{b :(b,a) \in attacks} (d(a) \vee d(b)))$;
however this part is logically equivalent to the first part of
$\alpha(AF)$ which is $(\bigwedge_{b :(b,a) \in attacks} d(a)
\leftarrow \neg d(b))$. In fact, the main difference is their
behavior \wrt stable model semantics. In order to illustrate this
difference, let us consider the argumentation framework $AF =
\langle \{ a \},\{(a,a)\} \rangle$. We can see that
$$\Gamma_{AF} = \{ d(a) \vee d(a) \} \cup \{ d(a) \leftarrow d(a) \}$$
and $$\alpha(AF) = (d(a) \leftarrow \neg d(a)) \wedge (d(a)
\leftarrow d(a))$$

\noindent It is clear that both formul\ae\ have a minimal model
which is $\{d(a)\}$\footnote{Notice that $\{d(a)\}$ suggests that
$AF$ has a preferred extensions which is $\{\}$.}; however
$\alpha(AF)$ has no stable models. This suggests that $\alpha(AF)$
is not a suitable representation for characterizing preferred
extensions by using stable models. Nonetheless we will see that the
stable models of $\Gamma_{AF}$ characterize the preferred extensions
of $AF$.


Even though, in this paper we are only interested in the preferred
semantics, it is worth mentioning that the stable models of the
first part of the formula $\alpha(AF)$ \ie $(\bigwedge_{b :(b,a) \in
attacks} d(a) \leftarrow \neg d(b))$, characterize the so called
stable semantics in argumentation theory \cite{Dung95}. It is also
important to point out that $\alpha(AF)$ and $\Gamma_{AF}$ have
different use. On the one hand, we will see that $\Gamma_{AF}$ is a
suitable mapping for inferring preferred extensions by using stable
model solvers. On the other hand, $\alpha(AF)$ has shown to be most
suitable for studying abstract argumentation semantics. For example
in \cite{NivOsoCorOG06}, $\alpha(AF)$ was used for defining an
extension of the preferred semantics. Also, since the well-founded
model of $\alpha(AF)$ characterizes the grounded semantics of $AF$,
$\alpha(AF)$ was used for defining extensions of the grounded
semantics and to describe the interaction of arguments based on
reasoning under the grounded semantics \cite{TR:NivOsoCor08}.

\end{remark}

In the following theorem we formalize a characterization of the
preferred semantics in terms of positive disjunctive logic programs
and stable model semantics.

\begin{theorem}\label{prefe-General}
Let  $AF = \langle AR,attacks\rangle$ be an argumentation framework
and $S \subseteq AR$. $S$ is a preferred extension of $AF$ if and
only if $compl(S)$ is a stable model of $\Gamma_{AF}$.
\end{theorem}
\begin{proof}
See Appendix A.

\end{proof}

Let us consider the following example.

\begin{figure}[th]\label{}
\begin{center}
\epsfig{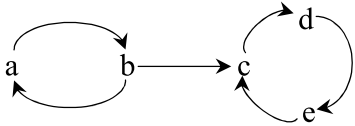} \caption{Graph representation of the argumentation
framework $AF  = \langle \{a,b,c,d,e\},
\{(a,b),(b,a),(b,c),(c,d),(d,e),(e,c)\}$}.\label{fig2}
\end{center}
\end{figure}

\begin{example}\label{prefer-general-prog}
Let $AF$ be the argumentation framework of \fig \ref{fig2}. We can
see that $\Gamma_{AF}$ is:

\begin{center}
\begin{tabular}{ll}
  $d(a) \vee d(b).$ $~~~~~~~~~~~~~~~~~$ & $d(a) \leftarrow d(a).$ \\
  $d(b) \vee d(a).$ & $d(b) \leftarrow d(b).$ \\
  $d(c) \vee d(b).$ & $d(c) \vee d(e).$ \\
  $d(c) \leftarrow d(a).$ & $d(c) \leftarrow d(d).$ \\
  $d(d) \vee d(c).$ & $d(d) \leftarrow d(b),d(e).$ \\
  $d(e) \vee d(d).$ & $d(e) \leftarrow d(c).$ \\
\end{tabular}
\end{center}

\noindent $\Gamma_{AF}$ has two stable models which are
$\{d(a),d(c),d(e) \}$ and $\{d(b),d(c),d(e),d(d))\}$, therefore
$\{b,d\}$ and $\{a\}$ are the preferred extensions of AF.
\end{example}

\subsection{Default negation}

As we have commented in whole paper, ours mappings are inspired by
two basic conditions that make an argument to be defeated. One of
the advantages of characterizing the preferred semantics by using a
logic programming semantics with \emph{default negation}, is that we
can infer the acceptable arguments from the stable models of
$\Gamma_{AF}$ in a straightforward form. For instance, let
$\Lambda_{AF}$  be the disjunctive logic program $\Gamma_{AF}$ of
Example \ref{prefer-general-prog} plus the following clauses:

\begin{center}
\begin{tabular}{ll}
  $ a \leftarrow \neg d(a).$ & $ b \leftarrow \neg d(b).$\\
  $ c \leftarrow \neg d(c).$ & $ d \leftarrow \neg d(d).$\\
  $ e \leftarrow \neg d(e).$ & \\
\end{tabular}
\end{center}

\noindent such that the intended meaning of each clause is: the
argument $x$ is acceptable if it is not defeated. $\Lambda_{AF}$ has
two stable models which are $\{ d(a), d(c), d(e), b, d \}$ and
$\{d(b), d(c), d(e), d(d), a \}$. By taking the intersection of each
model of $\Lambda_{AF}$ with $AR$ (the set of arguments of $AF$), we
can see that $\{b ,d\}$ and $\{a\}$ are the preferred extensions of
$AF$. This idea is formalized by Proposition \ref{lemma:prefDisjNF}
below.

\begin{definition}
Let  $AF = \langle AR,attacks\rangle$ be an argumentation framework.
We define its associated general program as follows:

\[
\Lambda_{AF} = \bigcup_{a \in AR}  \{ \Gamma(a) \cup  \{a \leftarrow
\neg d(a)\} \}
\]

\end{definition}

Notice that $\Gamma(a)$  and $\Lambda(a)$ are equivalent, the main
difference between $\Gamma_{AF}$ and $\Lambda_{AF}$ is the rule $a
\leftarrow \neg d(a)$ for each argument.

\begin{proposition}\label{lemma:prefDisjNF}
Let  $AF = \langle AR,attacks\rangle$ be an argumentation framework
and $S \subseteq AR$. $S$ is a preferred extension of $AF$ iff there
is a stable model $M$ of $\Lambda_{AF}$ such that $S = M \cap AR$.
\end{proposition}
\begin{proof}
The proof is straightforward from Theorem \ref{prefe-General} and
the semantics of default negation.
\end{proof}

It is worth mentioning that by using the disjunctive logic program
$\Lambda_{AF}$  and the DLV System, we can perform any query \wrt
\emph{sceptical and credulous reasoning}. For instance let
\texttt{gamma-AF} be the file which contains $\Lambda_{AF}$ such
that $AF$ is the argumentation framework of Fig. \ref{fig2}. Let us
suppose we want to know if the argument \emph{a} belongs to some
preferred extension of $AF$. Hence, let \texttt{query-1} be the
file:

$a?$

\noindent Let us call DLV with the \emph{brave/credulous reasoning}
front-end and \texttt{query-1}:

\noindent \texttt{\$ dlv -brave gamma-AF query-1}

\noindent \emph{a} \texttt{is bravely true, evidenced by} $\{d(b),
d(c), d(e), d(d), a \}$

\noindent This means that it is true that the argument $a$ belongs
to a preferred extension and even more we have a preferred extension
which contains the argument $a$. Now let us suppose that we want to
know if the argument $a$ belongs to all the preferred extensions of
$AF$. Let us call DLV with the \emph{cautious/sceptical reasoning}
front-end and \texttt{query-1}:

\noindent \texttt{\$ dlv -cautious gamma-AF query-1}

\noindent $a$ \texttt{is cautiously false, evidenced by} $\{ d(a),
d(c), d(e), b, d \}$

\noindent This means that it is false that the argument $a$ belongs
to all the preferred extensions of $AF$. In fact, we have a
counterexample.

%% file: proofs.tex
\section*{Appendix A}

\subsection*{Proof of Proposition \ref{pseudo-dualidad}}
\begin{proof}
First of all two observations:

\begin{enumerate}
  \item Given $M_1, M_2 \subseteq {\cal L}_T$, it is true that
  $M_1 \subset M_2$ iff $f({\cal L}_T \setminus M_2)
\subset f({\cal L}_T \setminus M_1)$.
  \item Given a propositional formula $A$, an interpretation $M$ from
  ${\cal L}_T$ to $\{0,1\}$ and $x \in \{0,1\}$. Then it is not
  difficult to prove by induction on $A$'s length\footnote{Since $A$ is a disjunctive clause,
  the length of $A$ is given by the number of atoms in the head of $A$ plus the number of literals in the body of $A$.}
  that  $M(A)= x$ iff $f({\cal L}_T \setminus M)(g(A)) = x$.

\end{enumerate}

\begin{description}
\item[=>]  To prove that if $M$ is a
maximal model of $T$ then $f({\cal L}_T \setminus M)$ is a minimal
model of $g(T)$. The proof is by contradiction. Let us suppose that
$M$ is a maximal model of $T$ but $f({\cal L}_T \setminus M)$ is a
model of $g(T)$ and is not minimal. Then if $f({\cal L}_T \setminus
M)$ is not minimal then there exists $M_2$ such that $f({\cal L}_T
\setminus M_2)$ is a model of $g(T)$ and $f({\cal L}_T \setminus
M_2) \subset f({\cal L}_T \setminus M)$. Then by observation 2, if
$f({\cal L}_T \setminus M_2)$ is a model of $g(T)$ then $M_2$ is a
model of $T$. By observation 1, if $f({\cal L}_T \setminus M_2)
\subset f({\cal L}_T \setminus M)$ then $M \subset M_2$. But this is
a contradiction because $M$ is a maximal model of $T$.
\item[<=] 

To prove that if $f({\cal L}_T \setminus M)$ is a minimal model of
$g(T)$ then $M$ is a maximal model of $T$. The proof is also by
contradiction. Let us suppose that $f({\cal L}_T \setminus M)$ is a
minimal model of $g(T)$ but $M$ is model of $T$ and is not maximal.
If $M$ is not maximal, then exists a model $M_2$ of $T$ such that $M
\subset M_2$. Then by observation 2, if $M_2$ is a model of $T$ then
$f({\cal L}_T \setminus M_2)$ is a model of $g(T)$. By observation
1, if $M \subset M_2$ then $f({\cal L}_T \setminus M_2) \subset
f({\cal L}_T \setminus M)$. But this is a contradiction because
$f({\cal L}_T \setminus M)$ is a minimal model of $g(T)$.

\end{description}
\end{proof}

\subsection*{Proof of Theorem \ref{theo:mimModelPref}}
\begin{proof}
Two observations:

\begin{enumerate}
  \item Since  the mapping $f(x)$ corresponds
to $d(x)$, then $compl(S) = f(AR\setminus S)$ because $compl(S) =
\{d(a) | a \in AR\setminus S\}$ and $f(AR\setminus S) = \{f(a) |
a\in AR\setminus S\}$.
  \item $\alpha(AF)$ is logically equivalent to $g(\beta(AF))$:\\

$g(\beta(AF)) =$
$$
\bigwedge_{a \in AR} ((\neg d(a) \rightarrow \bigwedge_{b :(b,a) \in
attacks} d(b)) \wedge (\neg d(a) \rightarrow \bigwedge_{b :(b,a) \in
attacks} ( \;\;\;\; \bigvee_{c :(c, b) \in attacks} \neg d(c) )))$$

\noindent Since $a \rightarrow \bigwedge_{b \in S} b \equiv
\bigwedge_{b \in S}(a \rightarrow b)$, we get:

$$ \bigwedge_{a \in AR} (\bigwedge_{b :(b,a) \in attacks}(\neg d(a)
\rightarrow d(b)) \wedge (\bigwedge_{b :(b,a) \in attacks} (\neg
d(a) \rightarrow  \bigvee_{c :(c,b) \in attacks} \neg d(c) )))$$

\noindent By applying transposition and cancelation of double
negation in both implications, we get:

$$\bigwedge_{a \in AR} (\bigwedge_{b :(b,a) \in attacks}(\neg d(b)
\rightarrow d(a)) \wedge (\bigwedge_{b :(b,a) \in attacks} ( \neg
\bigvee_{c :(c,b) \in attacks} \neg d(c) \rightarrow d(a) )))$$

\noindent Now, for the right hand side of the formula we need to
apply Morgan laws:

$$\bigwedge_{a \in AR} (\bigwedge_{b :(b,a) \in attacks}(\neg d(b)
\rightarrow d(a)) \wedge (\bigwedge_{b :(b,a) \in attacks} (
\bigwedge_{c :(c,b) \in attacks}  d(c) \rightarrow d(a) )))$$

\noindent Finally by changing $\rightarrow$ by $\leftarrow$, we get
$\alpha(AF)$.

$$\bigwedge_{a \in AR} (\bigwedge_{b :(b,a) \in attacks}( d(a)
\leftarrow \neg d(b)) \wedge ( \bigwedge_{b :(b,a) \in attacks} (
d(a) \leftarrow \bigwedge_{c :(c,b) \in attacks}  d(c) ))) = $$

\noindent $\alpha(AF)$

\end{enumerate}

\noindent Now the main proof: $S$ is a preferred extension of $AF$
iff (by Proposition \ref{propBesDou04}) $S$ is a maximal model of
$\beta(AF)$ iff (by Proposition \ref{pseudo-dualidad})
$f(AR\setminus S)$ is a minimal model of $g(\beta(AF))$ iff (by
observations 1 and 2) $compl(S)$ is a minimal model of $\alpha(AF)$.


\end{proof}

\subsection*{Proof of Proposition \ref{lemmaConsequence}}

First of all, let us introduce the following relationship between
minimal models and logic consequence.

\begin{lemma}\cite{OsNaAr:tplp}\label{lemma:minMondelConsLogic}
For a given general program $P$, $M$ is a model of $P$ and $P \cup
\lnot\setcomp{M)} \models M$ iff $M$ is a minimal model of $P$.
\end{lemma}

This lemma was introduced in terms of augmented programs. Since a
general program is a particular case of an augmented program, we
write the lemma in terms of general programs (see \cite{OsNaAr:tplp}
for more details about augmented programs).

\begin{proof}
$S$ is a preferred extension of $AF$ iff (by Theorem
\ref{theo:mimModelPref} ) $compl(S)$ is a minimal model of
$\alpha(AF)$ iff (by lemma \ref{lemma:minMondelConsLogic})
$compl(S)$ is a model of $\alpha(AF)$ and $\alpha(AF) \wedge
SetToFormula( \lnot\setcomp{compl(S))} \models
SetToFormula(compl(S))$.
\end{proof}

\subsection*{Proof of Theorem \ref{prefe-General}}
\begin{proof}
\emph{S} is a preferred extension of \emph{AF} iff \emph{compl(S)}
is a minimal model of $\alpha(AF)$ (by Theorem
\ref{theo:mimModelPref}) iff $compl(S)$ is a minimal model of
$\Gamma_{AF}$ (since $\Gamma_{AF}$ is logically equivalent to
$\alpha(AF)$ in classical logic)  iff \emph{compl(S)} is a stable
model of $\Gamma_{AF}$ (since $\Gamma_{AF}$ is a positive
disjunctive logic program and for every positive disjunctive logic
program \emph{P}, \emph{M} is a stable model of \emph{P} iff
\emph{M} is a minimal model of \emph{P}).
\end{proof}